\documentclass[conference]{IEEEtran}
\usepackage[T1]{fontenc}
\IEEEoverridecommandlockouts
\usepackage{cite}
\usepackage{amsmath,amssymb,amsfonts}
\usepackage{algorithmic}
\usepackage{graphicx}
\usepackage{textcomp}
\usepackage{xcolor}
\usepackage[utf8]{inputenc}
\usepackage{textgreek}
\usepackage{subcaption}
\def\BibTeX{{\rm B\kern-.05em{\sc i\kern-.025em b}\kern-.08em
    T\kern-.1667em\lower.7ex\hbox{E}\kern-.125emX}}

\makeatletter
\newcommand{\linebreakand}{%
    \end{@IEEEauthorhalign}
    \hfill\mbox{}\par
    \mbox{}\hfill\begin{@IEEEauthorhalign}
}

\begin{document}
\title{Implementing a \textit{Sharia} Chatbot as a Consultation Medium for Questions About Islam\\}

\author{\IEEEauthorblockN{1\textsuperscript{st} Wisnu Uriawan}
\IEEEauthorblockA{\textit{Informatics Department}\\
\textit{UIN Sunan Gunung Djati Bandung}\\
West Java, Indonesia\\
wisnu\_u@uinsgd.ac.id}
\and
\IEEEauthorblockN{2\textsuperscript{nd} Aria Octavian Hamza}
\IEEEauthorblockA{\textit{Informatics Department}\\
\textit{UIN Sunan Gunung Djati Bandung}\\
West Java, Indonesia\\
ariaoctavianhamza@gmail.com}
\and
\IEEEauthorblockN{3\textsuperscript{rd} Ade Ripaldi Nuralim}
\IEEEauthorblockA{\textit{Informatics Department}\\
\textit{UIN Sunan Gunung Djati Bandung}\\
West Java, Indonesia\\
ripaldinuralim@gmail.com}
\linebreakand
\IEEEauthorblockN{4\textsuperscript{th} Adi Purnama}
\IEEEauthorblockA{\textit{Informatics Department}\\
\textit{UIN Sunan Gunung Djati Bandung}\\
West Java, Indonesia\\
adipurnamaa8@gmail.com}
\and
\IEEEauthorblockN{5\textsuperscript{th} Ahmad Juaeni Yunus}
\IEEEauthorblockA{\textit{Informatics Department}\\
\textit{UIN Sunan Gunung Djati Bandung}\\
West Java, Indonesia\\
zen.informatika9@gmail.com}
\and
\IEEEauthorblockN{6\textsuperscript{th} Anissya Auliani Supriadi Putri}
\IEEEauthorblockA{\textit{Informatics Department}\\
\textit{UIN Sunan Gunung Djati Bandung}\\
West Java, Indonesia\\
anissyaauliani@gmail.com}
}
\maketitle

\begin{abstract}
This research presents the implementation of a \textit{Sharia}-compliant chatbot as an interactive medium for consulting Islamic questions, leveraging Reinforcement Learning (Q-Learning) integrated with Sentence-Transformers for semantic embedding to ensure contextual and accurate responses. Utilizing the CRISP-DM methodology, the system processes a curated Islam QA dataset of 25,000 question-answer pairs from authentic sources like the \textit{Qur’an}, \textit{Hadith}, and scholarly \textit{fatwas}, formatted in JSON for flexibility and scalability. The chatbot prototype, developed with a Flask API backend and Flutter-based mobile frontend, achieves 87\% semantic accuracy in functional testing across diverse topics including \textit{fiqh}, \textit{aqidah}, \textit{ibadah}, and \textit{muamalah}, demonstrating its potential to enhance religious literacy, digital da’wah, and access to verified Islamic knowledge in the Industry 4.0 era. While effective for closed-domain queries, limitations such as static learning and dataset dependency highlight opportunities for future enhancements like continuous adaptation and multi-turn conversation support, positioning this innovation as a bridge between traditional Islamic scholarship and modern AI-driven consultation.
\end{abstract}


\section{Introduction} \label{sec:introduction}

The rapid development of digital technology over the last two decades has brought significant changes to various aspects of human life, including the way people interact, learn, and obtain information. One crucial breakthrough in the Industry 4.0 era is the advancement in the fields of Artificial Intelligence (AI) and Natural Language Processing (NLP). These two fields enable machines to understand, interpret, and respond to human language in a smart and contextual manner. This innovation has given birth to various human-computer interaction applications, one of which is the chatbot.

A chatbot is an AI-based system designed to simulate human-like conversations through text or voice. The chatbot functions as a virtual assistant that can automatically provide responses to user questions. This technology is now widely implemented across various sectors, ranging from customer service, education, health, finance, to public services. The use of chatbots has been proven to increase service efficiency, accelerate access to information, and minimize the limitations of human time and effort\cite{uriawan2024real}.

In the context of education and religious guidance, particularly Islam, the development of information technology presents both opportunities and new challenges. Muslims in the digital age face a massive flood of information, not all of which has validity and adherence to the principles of Islamic \textit{Sharia}. Many people, especially the younger generation, seek quick answers about Islamic laws, worship procedures, and social religious issues through the internet. However, the limitations of trustworthy sources and the variation in interpretations often lead to confusion and the potential for the spread of inaccurate information \cite{anggraini2024islamic}.

Therefore, there is a need for an AI-based digital consultation system that can answer questions about Islam quickly, accurately, and in accordance with \textit{Sharia} principles. One promising approach to address this need is the implementation of a \textit{Sharia} chatbot, which is an NLP-based system specifically trained using a dataset containing Islamic material such as verses from the \textit{Qur'an}, \textit{Hadith}, scholarly fa\textit{}twas, and interpretive and \textit{Fiqh }(Islamic jurisprudence) literature. This chatbot can act as an interactive religious consultation medium, helping users understand Islamic teachings easily, structurally, and contextually \cite{masuzzahra2025hana}.

In its development, the \textit{Sharia} chatbot system requires effective NLP modeling to recognize the intent of user questions (intent detection) and provide relevant answers (response generation). To achieve this, a high-quality, systematically compiled dataset is needed. The JSON (JavaScript Object Notation) format is an ideal choice due to its light weight, flexibility, and easy integration with machine learning algorithms. Each entry in the JSON dataset contains a question-answer pair, which can be expanded with additional information such as references to verses, \textit{Hadith}, or classical book sources, thereby increasing the system's credibility and scholarly context \cite{sofyan2025implementation}.

Furthermore, utilizing JSON as the dataset format facilitates developers in updating content, adding new topics, or expanding the scope of the knowledge domain in line with the dynamics of issues faced by the Muslim community. By employing \textit{(machine learning)} or deep learning approaches, the chatbot can continuously adapt and improve its response accuracy based on user interactions \cite{campbell2020authority}.

Beyond the technical aspects, the presence of the \textit{Sharia} chatbot also holds significant social and digital da'wa (preaching/invitation to Islam) dimensions. In the era of social media and the digitalization of da'wa, Muslims require a medium that is not only informative but also interactive and educational. An Islamic Chatbot can function as a modern means of da'wa, integrating \textit{Sharia} values with advancements in information technology. This system has the potential to become an intermediary between scholars and the community, especially for those who struggle to access conventional sources of knowledge \cite{okonkwo2021chatbots}.

Empirically, various previous studies have shown that the application of NLP-based chatbots in education and consultation can increase the effectiveness of information delivery, strengthen user engagement, and foster self-learning interest. In the context of Islam, this aligns with the spirit of Iqra' the first command in the Qur'an which encourages the community to continuously learn and seek knowledge in a wise and civilized manner \cite{zulfa2025peran}.

Thus, the implementation of a \textit{Sharia} chatbot as a consultation medium for questions about Islam does not only represent the integration between technology and religion, but also constitutes a strategic innovation in supporting digital transformation in the fields of education and da'wa. This chatbot is expected to become a smart religious assistant that bridges the gap between the community's need for information and the limited access to authentic sources of knowledge \cite{hardiyanti2025pengembangan}.

By leveraging the power of NLP and JSON-based datasets, this system has the potential for sustainable development, both locally and globally, to support the vision of Islamic Digital Transformation. Moving forward, the development of the \textit{Sharia} chatbot can also be directed towards supporting service personalization, voice integration (speech recognition), and more in-depth semantic analysis to make the interaction between users and the system increasingly natural, humanistic, and educationally valuable.

\section{Related Work} \label{sec:related-work}

The development of information technology has driven a significant transformation in the application of chatbots, particularly within the context of Islamic religion. Initially, developed chatbot systems were closed-domain, meaning they could only answer specific questions based on a static knowledge base. For instance, a study developed a chatbot to answer questions about salat (prayer) and \textit{zakat} using the Fuzzy String-Matching algorithm \cite{Sihotang2020}. While effective for limited scenarios, this system had limitations in understanding the variations of natural language and more complex question contexts. At a global level, a hybrid architecture was used to enhance the chatbot's ability to handle Arabic-language Islamic legal questions \cite{Othman2022}.

With the advancement of technology, more sophisticated approaches using deep learning, especially Convolutional Neural Networks (CNN), were introduced. Research implemented a CNN-based chatbot for an Islamic question-and-answer system, capable of providing automatic answers based on user commands in natural language \cite{Suwarman2025}. This approach, which utilized transfer learning and word embeddings techniques, proved to increase the system's ability to understand more complex question contexts. Significant transformation has also been observed with the adoption of Large Language Models (LLMs) such as GPT-3 and Gemini AI. A study used a LangChain-based chatbot for \textit{fiqh} \textit{muamalah} (Islamic transaction law) topics, with an answer accuracy reaching 88.8\% \cite{Nurhapiza2024}. Another work developed a chatbot system using Gemini-2.0-flash integrated with LangChain to answer contemporary \textit{fiqh} questions based on the works of Shaykh Al-Qardhawi, with evaluations showing the satisfaction level of \textit{fiqh} experts reaching 89\% \cite{Helviansyah2025}.

The validity of information sources is a critical factor in the development of \textit{Sharia} chatbots. One research emphasized the importance of using authentic sources of \textit{fiqh} from the four schools of thought (madzhab) \cite{Rahayu2024}, while others applied LLM and LangChain to build a chatbot capable of answering questions related to \textit{Tafsir} \textit{Al-Azhar} via the Telegram platform \cite{Permadi2024}. Validating these sources ensures that the answers provided comply with \textit{Sharia} principles and can be relied upon as a consultation medium.

In addition to technical aspects, ethics and user acceptance are important considerations. Studies asserted that AI can only provide basic information regarding digital family law and cannot replace human consultation \cite{Falah2023}. Consistent with this finding, other work highlighted the risk of depersonalization in the use of AI chatbots for religious consultation, emphasizing that human interaction remains necessary to maintain the social and humanitarian dimension \cite{Insana2024}. A case study on the AISYAH BSI Chatbot demonstrated that usability and responsiveness are key determinants of user satisfaction with digital \textit{Sharia} services \cite{Fathir2024}.

From this literature review, a research gap is identified. Most studies focus on specific \textit{fiqh} domains and technical metrics such as answer accuracy, without a systematic evaluation of the chatbot's effectiveness as a comprehensive \textit{Sharia} consultation medium. This study aims to bridge this gap by designing and implementing a closed-domain \textit{Sharia} chatbot that covers a broader range of basic Islamic questions, while also analyzing user perception and its effectiveness. Thus, the resulting model is expected to be balanced between AI innovation and \textit{Sharia} compliance, yielding a holistic and user-acceptable digital consultation medium.

In recent years, research on \textit{Sharia} chatbots has shown a significant increase in both technology and the scope of religious domains. Early studies focused on developing an Android-based Islamic consultation chatbot application that emphasizes ease of access and integration with \textit{Hadith} and \textit{Qur'an} sources \cite{awaliyah2022chatbot}. Similar work highlighted the development of an AI-based Hajj and Umrah consultation chatbot to assist pilgrims in understanding the rituals and procedures of the pilgrimage journey \cite{hardiyanti2025haji}. Meanwhile, a framework for the ethics of a Halal chatbot (HalalBot) was proposed, emphasizing the principles of reliability, transparency, and compliance with Islamic law in providing halal certification information.

Natural Language Processing (NLP) approaches are also widely used to strengthen the semantic accuracy of Islamic chatbot systems. A study demonstrated the effectiveness of NLP in detecting the context of \textit{fiqh} questions and answers using the word2vec and BERT models . The results showed a significant increase in understanding the meaning of Indonesian utterances containing \textit{Sharia} terms. Other research also affirmed the importance of applying NLP to structure \textit{fiqh} datasets so that the system can provide answers that are not only relevant but also consistent with authentic \textit{fatwas} \cite{awaliyah2025syariah}.

In the context of digital transformation, several studies underscored the role of chatbots as a medium for da'wa (preaching) and religious literacy. Research stated that AI chatbots have great potential as an interactive digital da'wa strategy, allowing the wider dissemination of Islamic values through online media. Other work integrated chatbots into Islamic Religious Education learning in secondary schools, and the results showed a 23\% increase in student understanding . This study confirms that chatbots can serve a dual function, not only as a consultation tool but also as an adaptive, technology-based learning medium.

The ethical aspects and \textit{fiqh} limitations in the implementation of AI chatbots have been a concern for a number of researchers. Studies highlighted the phenomenon of AI globalization in Islamic law and warned of the need for new \textit{fatwas}\textit{fatwas} to regulate human interaction with automated systems. Ethical guidelines for the use of AI in the Islamic context were also comprehensively discussed, emphasizing that chatbots must be developed within the framework of maqasid \textit{sharia} (objectives of \textit{Sharia}) to preserve public interest \textit{(maslahat)} and avoid harm \textit{(mudharat)} \cite{ethical2024guidelines}. In line with this, other research highlighted the risk of dehumanization and the importance of religious moderation in utilizing AI chatbots \cite{Insana2024}.

Several government and religious institutional initiatives also reinforce this research direction. The Si PAHAM Project from the Indonesian Ministry of Religious Affairs is an example of implementing a religious moderation chatbot to early detect potential social conflicts and strengthen inter-religious tolerance \cite{sipaham2025sosialisasi}. This system demonstrates how AI can be adapted for broad social-religious purposes. Meanwhile, a Saudi Arabian technology company developed an Arabic-language chatbot aligned with Islamic values to expand global access to religious knowledge \cite{saudi2025chatbot}.

Contemporary research also shows new directions in the utilization of LLM and multimodal integration. Research introduced the HANA chatbot focused on \textit{fiqh} consultation for female menstruation based on GPT-3, with a user satisfaction level reaching 91\% . This study demonstrated that large language models are capable of handling highly specific domains of knowledge when supported by truly curated datasets. In addition, a study proposed the development of a smart chatbot as an interactive learning medium for Al-Islam based on adaptive semantic feedback, strengthening user engagement in the context of digital religious education.

The current research trend also shows a shift from rule-based systems to the hybrid-AI framework approach \cite{solomon2024rule}. Studies applied a CNN model combined with LLM in a contemporary \textit{fiqh} question-and-answer system \cite{ helviansyah2025kontemporer}, while others utilized LangChain integration to bridge semantic search and answer generation based on Qur'anic text (nash) \cite{Permadi2024}. This innovation marks the evolution of Islamic chatbots from mere static FAQ systems to intelligent entities capable of understanding contextual meaning and providing verifiable answers based on authoritative texts.

Overall, the current literature indicates that the main challenge in developing \textit{Sharia} chatbots is not only in the technical aspects but also in the epistemological and ethical aspects. The integration of AI in da'wa needs to consider the validity of the chain of scholarly transmission and the authority of scholars as guardians of the authenticity of Islamic law. Therefore, this research seeks to enrich the approach by balancing technological innovation, information accuracy, and adherence to \textit{Sharia} principles. This approach aligns with the vision of Islamic Digital Transformation, where AI is not just a tool but also a means to expand access to beneficial and ethical knowledge.

\section{Methodology} \label{sec:methodology}

Based on a poll conducted by datascience-pm.com in 2020, the CRISP-DM (Cross-Industry Standard Process for Data Mining) methodology was recorded as the most commonly used approach in data science projects, with an adoption rate reaching 49\%. This percentage is significantly higher than other methodologies such as Scrum (18\%), Kanban (12\%), and individual approaches (12\%). This dominance indicates that CRISP-DM remains the primary choice among data science practitioners because it offers a systematic, flexible, and easily adaptable workflow for various types of data analysis projects. Meanwhile, other methodologies such as TDSP (Team Data Science Process), SEMMA (Sample, Explore, Modify, Model, Assess), and other specific approaches garnered a relatively small proportion of usage. This suggests that explicit stage-based approaches like CRISP-DM are still more trusted in producing optimal analytical and data modeling processes across various industry sectors.
\begin{figure}[ht]
    \centering
    \includegraphics[width=\linewidth]{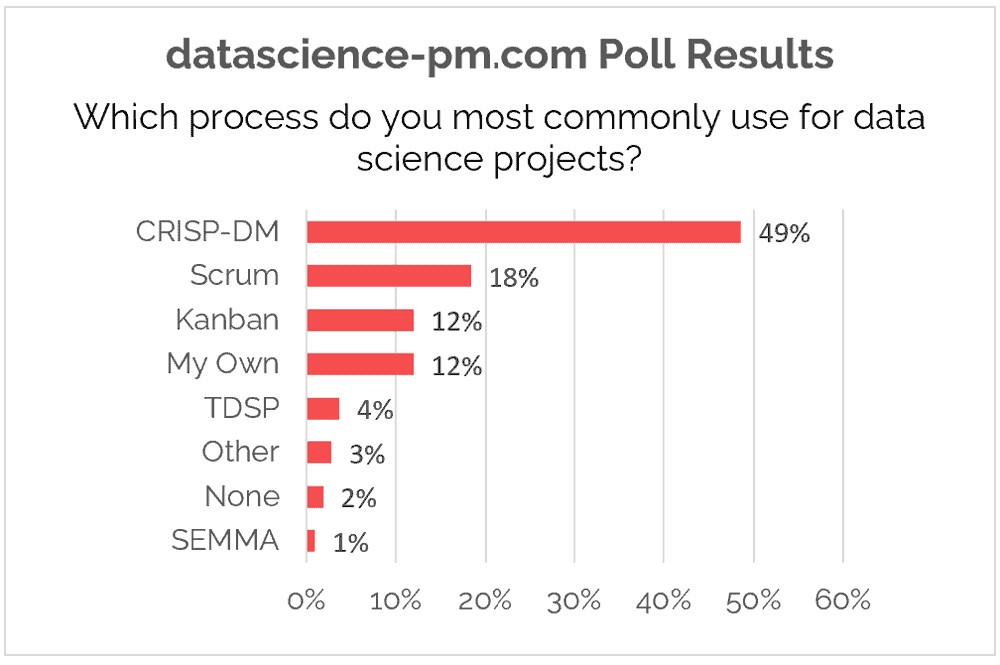}
    \caption{Popular survey methods}
    \cite{datasciencePM}
    \label{fig:metod}
\end{figure}

in line with this, this research utilizes the CRISP-DM (Cross Industry Standard Process for Data Mining) methodology as the main framework for the development of the \textit{Sharia} Chatbot application. CRISP-DM was chosen because it is capable of facilitating the system development process systematically and comprehensively, starting from problem understanding, data collection and preparation, modeling, evaluation, and implementation. This approach is highly relevant for research based on machine learning and natural language processing (NLP), as it ensures that every stage is carried out measurably and oriented towards improving model quality. By implementing CRISP-DM, the development process of the \textit{Sharia} chatbot can be more structured, allowing researchers to evaluate system performance comprehensively and produce a chatbot that is accurate, informative, and compliant with \textit{Sharia} principles.
\begin{figure}[htb!]
    \centering
    \includegraphics[width=1\linewidth]{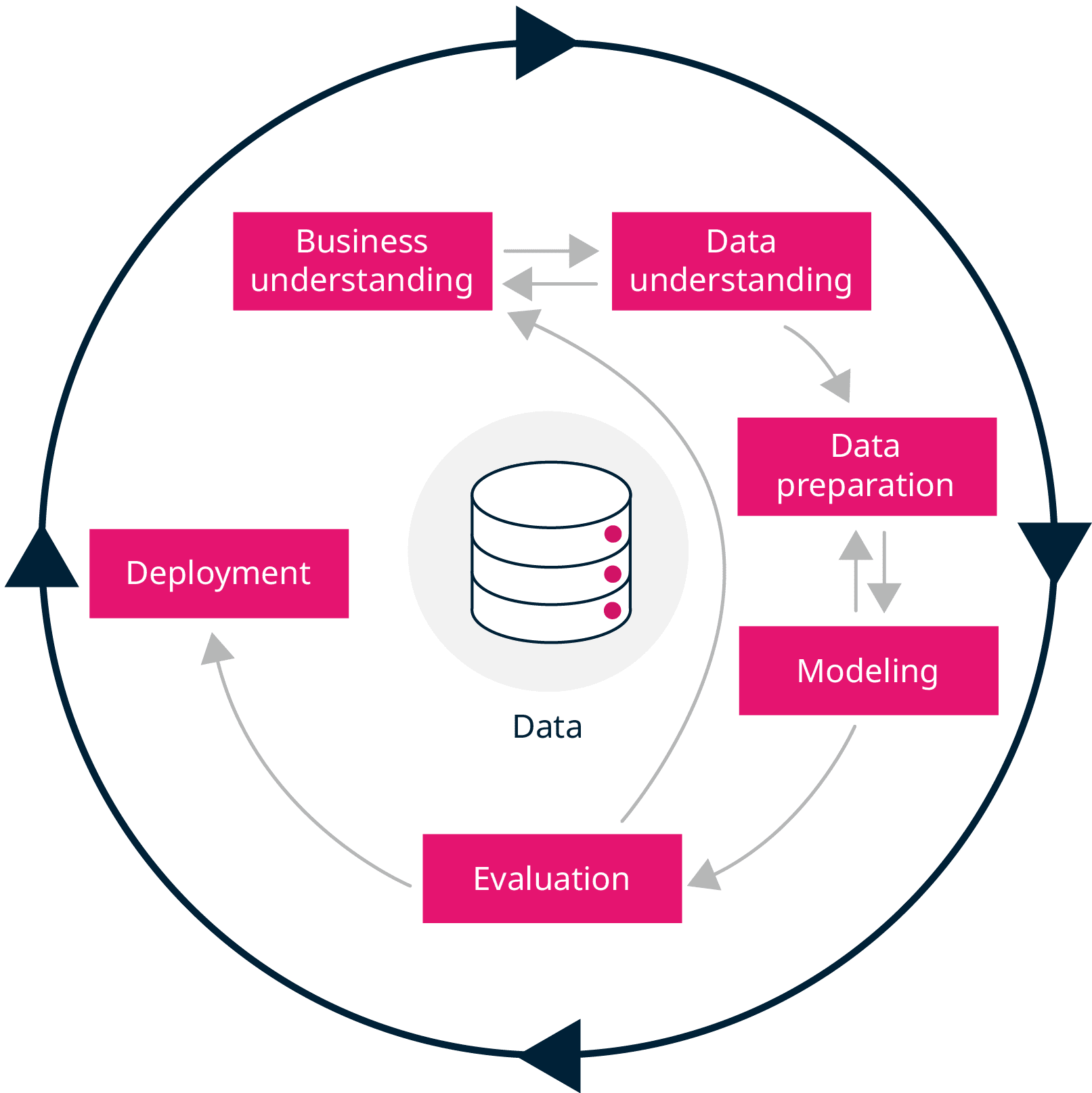}
    \caption{CRISP-DM Methodology}
    \cite{getsmarterdataMining}
    \label{fig:metod}
\end{figure}
CRISP-DM consists of six main phases used to manage machine learning-based application development projects efficiently and structurally. The following is the adjustment of each phase based on the focus of this research:

\subsection{Busines Understanding}
The first stage of the CRISP-DM methodology is Business Understanding, which focuses on a comprehensive understanding of the primary research objectives, user needs, and the benefits to be achieved through the development of a \textit{Sharia} chatbot system. This stage is crucial because it serves as the basis for determining the direction of system development, the scope of the problem, and the project's success criteria.

In the social and religious context, Muslims in the digital age face a major challenge in the form of a flood of information that cannot always be verified for accuracy. Many internet users seek quick answers about Islamic law, etiquette, and worship practices through online platforms such as forums, social media, or unofficial websites. However, many of these sources lack scientific authority and sometimes spread misinterpretations.
This problem highlights a gap in access to authentic and verified Islamic information, particularly among younger generations who interact more frequently through digital devices.

From a user needs perspective, the \textit{Sharia} chatbot is designed to be an interactive digital assistant that can answer Islamic questions quickly, contextually, and in accordance with \textit{Sharia }principles. This system can assist people seeking religious guidance without having to wait for in-person consultations with religious teachers or \textit{fatwa}\textit{fatwa} institutions. This chatbot also serves as an educational tool that can support self-learning in basic Islamic studies, such as \textit{fiqh} (Islamic jurisprudence), \textit{aqidah} (belief), morals, and daily worship.

Functionally, the development of a \textit{Sharia }Chatbot has several strategic objectives:
\begin{enumerate}
\item Enhancement of Information Accessibility
\item Efficiency in Religious Services
\item Digital Transformation of Da’wa
\item Validation of Authentic Islamic Knowledge
\item A Medium for Religious Literacy and Moderation
\end{enumerate}

From a business and technological perspective, the development of this chatbot also provides strategic added value for educational institutions and Islamic outreach organizations. With an AI-based system, religious institutions can expand the reach of Islamic outreach services without significantly increasing human resources. This model can be developed as a commercial API-based service (Chatbot-as-a-Service) that can be integrated into the internal platforms of Islamic boarding schools, Islamic universities, or public Islamic applications such as SiPAHAM from the Ministry of Religious Affairs.

With this approach, the system not only has social value but can also provide economic benefits and operational efficiency for the institutions that adopt it.

From a research perspective, this system also serves as a proof-of-concept prototype for the application of Reinforcement Learning (RL) in the domain of Islamic consultation. The use of Q-Learning with sentence transformers is expected to produce an adaptive and contextual chatbot, while also opening up opportunities for further research, such as reinforcement learning from human feedback (RLHF) or personalized answers based on user profiles \cite{hartawan2024bidirectional}.

Conceptually, the scope of system development includes several main aspects:
\begin{enumerate}
\item Domain Coverage: Basic Islamic questions covering fiqh, ethics \textit{akhlaq, aqidah, ibadah, and muamalah}.
\item System Limitations: The chatbot does not provide new legal \textit{fatwas} but only delivers answers based on existing authoritative sources.
\item Success Criteria: The system is considered successful if it can provide relevant answers with a minimum semantic accuracy of 85\%, a response time of less than 3 seconds, and a user satisfaction rate above 80\% based on User Acceptance Testing (UAT).
\end{enumerate}

In addition, this stage also produces the identification of the main stakeholders, namely:
\begin{enumerate}
\item General Users (End Users): The public who seek quick answers related to religious issues.
\item Academics and Researchers: Utilize the chatbot as a learning model or for \textit{Sharia}-based NLP research.
\item Religious Institutions: Adopt the system to expand the reach of digital da’wa.
\item Government and Regulators: Serve as a tool for religious literacy and moderation in the digital era.
\end{enumerate}

Thus, this business understanding phase emphasizes that the research aims not only to create an artificial intelligence-based system but also to support an ethical, measurable, and ummah-oriented digital transformation of Islam. The implementation of a \textit{Sharia} chatbot is expected to be a concrete solution for improving religious literacy, expanding humanistic da'wa (Islamic outreach), and strengthening Indonesia's position as a pioneer in Islamic digital technology in Southeast Asia.

\subsection{Data Understanding}
The Data Understanding stage aims to gain a deep understanding of the structure, quality, and characteristics of the data used in this research. Dataset quality plays a crucial role in determining the effectiveness of machine learning models, particularly for Natural Language Processing (NLP)-based systems like \textit{Sharia} chatbots.
This stage includes a descriptive analysis of the data sources, text length distribution, question context variations, and the relevance of the data content to the Islamic domain.

The main dataset used in this study is the Islam QA Dataset, obtained from Kaggle. This dataset was developed by the data science community with the goal of providing a corpus of Islamic question-and-answer texts sourced from international Islamic consultation websites such as \textit{IslamQA.info}, \textit{AboutIslam.net}, and several trusted online Islamic educational resources.
The dataset contains 32,000 question-answer pairs covering various Islamic topics, \textit{aqidah, ibadah, }fiqh\textit{, muamalah, akhlak, and tafsir}. Each entry in the dataset consists of two main components:

\begin{enumerate}
    \item Question: Questions asked by users or congregations regarding specific topics.
    \item Answer: Responses or explanations from \textit{Sharia} experts compiled based on sources from the Quran, hadith, and fatwas of scholars.
\end{enumerate}

In addition to these two main columns, some versions of the dataset also include additional metadata such as category (Islamic themes), reference links, and answer length, which can be used for further analysis.

This dataset is in CSV format with two main text columns and one additional category column. In general, the average length of questions is 14–20 words per entry, while the length of answers ranges from 80 to 250 words, depending on the complexity of the topic. The data distribution shows that the categories of \textit{fiqh} and worship dominate about 47\% of the total dataset, followed by \textit{muamalah} (23\%), \textit{aqidah} (15\%), \textit{akhlak} (10\%), and the rest are \textit{tafsir} and Islamic history (5\%).

This analysis is important for understanding the proportion and bias of themes, as an unbalanced distribution of categories can affect the chatbot's ability to answer certain types of questions.

\subsection{Data Preparation}
The Data Preparation stage is one of the most crucial processes in developing a Natural Language Processing (NLP)-based system, as the quality of the data used significantly determines the accuracy and reliability of the resulting model. After the Data Understanding stage is complete, the next step is to prepare the dataset for optimal use by machine learning algorithms, specifically Reinforcement Learning (Q-Learning) and sentence-transformer models.

In this study, the data preparation stage was conducted on the Islam QA Dataset, which had undergone a previous exploration process. The dataset contains question and answer pairs (Q and A pairs) sourced from various Islamic topics such as worship, \textit{fiqh} (Islamic jurisprudence), \textit{muamalah} (religious affairs), \textit{aqidah} (belief), and morals. The data preparation process was carried out carefully to ensure that each entry used was relevant, clean, and consistent in format and context.

The first step in the preparation stage was an initial data selection to ensure only entries that align with the system's requirements were included. The selection process was based on three main criteria:
\begin{enumerate}
\item Thematic Relevance: Questions that are directly related to Islamic topics (for example, worship law, ethics, or modern muamalah). Non-religious questions, extremely personal ones, or polemical questions outside the context of \textit{Sharia} are removed.
\item Completeness of Q and A Pairs: Only entries that have complete questions and answers are retained. Entries with empty answer columns or answers that are too short ($<$20 words) are deleted.
\item Text Cleanliness: Data containing foreign characters, HTML tags, or special symbols are cleaned before entering the preprocessing stage.
\end{enumerate}

After the filtering process, the usable data set was 25,000 question-answer pairs, with a (100\%)completeness rate across both main columns. Manual validation was performed on approximately 300 random samples to ensure that each answer was sourced from a valid Islamic context.

Next, text cleaning was performed. This stage aims to remove irrelevant elements and transform the entire text into a standard format for consistent processing by the NLP model.
The steps involved are as follows:
\begin{enumerate}
\item Removal of Special Characters: All non-alphabetic symbols such as numbers, double punctuation marks, and HTML formats ($<p>, <br>$) are removed using regular expressions (regex).
\item Letter Normalization: All text is converted to lowercase (lowercase transformation) to avoid duplication due to capitalization differences.
\item Stopword Removal: Common words such as the, and, is, as well as their Indonesian equivalents like yang, dan, itu, are removed so that the analysis focuses on contextually meaningful words.
\item Lemmatization and Stemming: Each word is returned to its base form. For English text, WordNetLemmatizer is used, while for Indonesian text, Sastrawi Stemmer is applied.
\item Semantic Duplication Removal: Questions that have similar meanings but different wording are identified through cosine similarity, and one of the most representative ones is selected.
\end{enumerate}

These steps produce a dataset with a clean, consistent sentence structure, ready for use in the tokenization and modeling process.

After the text is cleaned, tokenization is performed, which involves breaking sentences into word units (tokens). Tokenization is performed using the NLTK tokenizer for English and the IndoNLP tokenizer for Indonesian data.
This process is crucial for the model to recognize word boundaries, calculate word frequency, and facilitate the formation of semantic vectors in the next stage.
This step produces a corpus ready to be mapped into numeric form using sentence transformers. These tokens form the basis for the embedding process used by the RL agent to calculate semantic similarity between questions.

Because the Q-Learning model works optimally with relatively uniform text representations, answer length normalization is performed to avoid data imbalance.
Initial analysis results indicate that approximately (30\%) of the entries are longer than 250 words. To maintain conciseness yet informativeness, automatic extractive summarization is performed using a TF-IDF sentence ranking algorithm.
This process selects the two to three most representative sentences from the answers to be used as the chatbot's final response. This approach maintains the substance of the \textit{sharia} law without cutting out important meanings such as verses or \textit{hadith}. The Dataset are include:
\begin{enumerate}
    \item Training Data: 80\% of the total dataset ($\approx$20,000 entries) is used to build the Q-table and learn semantic similarity patterns.
    \item Testing Data: The remaining 20\% ($\approx$5,000 entries) is used to evaluate the chatbot's ability to answer new questions that it has not been trained on.
\end{enumerate}

The division was performed using stratified sampling to maintain a balanced topic distribution across both data subsets. Thus, the Data Preparation stage produces a structured, clean, and relevant dataset for the Islamic domain, ready for use in the embedding and Q-Learning processes. This process also ensures that the chatbot developed is not only technically intelligent but also adheres to the principles of source validity and the accuracy of religious information.

\subsection{Modeling}
At this stage, a Reinforcement Learning (RL)-based chatbot was developed using a Q-learning approach combined with semantic representations from the sentence-transformers model. The main objective of this model is to select the most relevant answer from the available data set based on the user's question, taking into account contextual semantic similarity.

In this context, the agent (chatbot) interacts with an environment consisting of a dataset of questions and answers. The representation of user questions, which become states in the Q-learning algorithm, is converted into numerical vectors using the sentence-transformers model (in this case, $paraphrase-multilingual-MiniLM-L12-v2$), which is capable of mapping sentences into fixed-dimensional embedding representations. These embedding vectors are then compared with all existing question embeddings in the dataset using a semantic similarity level (cosine similarity) to measure the semantic proximity between questions.

Each action in the system is defined as selecting one of the answers from the dataset. The selection process is based on the highest reward value calculated from the semantic similarity between the user's question and the original question from the available answers. The higher the semantic similarity value, the greater the reward given for that action. This reward can be a full value if the similarity exceeds a certain threshold (e.g., >0.8), a partial reward if it is in the middle range, or even a negative penalty if it is too low.

The Q table is constructed as a two-dimensional matrix that stores the Q value for each combination of states and actions. The Q value is updated iteratively using the Q-learning formula:

\begin{equation}
Q(s,a) \leftarrow Q(s,a) + \alpha \left[ r + \gamma \max_{a'} Q(s',a') - Q(s,a) \right]
\end{equation}\\

where α is the learning rate, γ is the discount factor, and r is the reward based on semantic similarity. With this approach, the system learns to associate specific questions with the most relevant answers based on patterns of semantic similarity.
\begin{figure}[htb!]
    \centering
    \begin{subfigure}[b]{0.25\columnwidth}
        \centering
        \includegraphics[width=\textwidth]{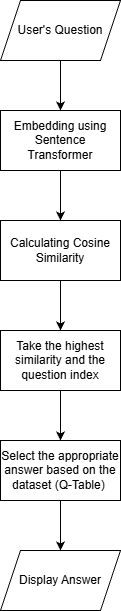}
        \caption{Flowchart of the Q-Learning based chatbot process}\label{3a}
        \label{fig:sub_a}
    \end{subfigure}
    \hspace{1cm}
    \begin{subfigure}[b]{0.25\columnwidth}
        \centering
        \includegraphics[width=\textwidth]{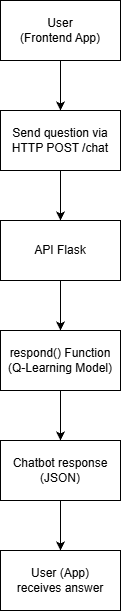}
        \caption{Flowchart of how the API works using FLASK}\label{3b}
        \label{fig:sub_b}
    \end{subfigure}
    \caption{Comparison of the Q-Learning–based chatbot workflow (a) and the API communication workflow using Flask (b) presented in a single column.}
    \label{fig:sidebyside}
\end{figure}

\subsection{API}
The next stage in this research is the development of an Application Programming Interface (API) as a bridge between the chatbot model and the front-end application that will be used by users. This API development aims to provide real-time, two-way communication between users and the chatbot.
The API was developed using Flask, a lightweight and flexible Python-based microweb framework. Flask was chosen for its several advantages, including:
\begin{enumerate}
    \item Lightweight and fast, making it suitable for simple API development with minimal resource requirements.   
    \item Easy to integrate with machine learning models, including the Q-Learning model developed in this research.
    \item Simple syntax that simplifies code implementation and maintenance.
\end{enumerate}

The API communication workflow can be seen in Figure \ref{3b} below:

\subsection{Mobile Application Prototyping}
As a follow-up to API development, this research also implemented a Flutter-based mobile application prototype that serves as the main user interface for direct interaction with the \textit{Sharia} chatbot. This application is designed to have a simple, lightweight, and responsive appearance on various Android and iOS devices. The purpose of developing this prototype is to validate system performance in a real environment, assess usability aspects, and ensure optimal data integration between the frontend and backend.

Flutter was chosen because it is an open-source framework developed by Google that supports cross-platform development. With a single codebase approach, Flutter enables the creation of a consistent user interface (UI) across different operating systems without the need to write separate code. Another advantage of Flutter is the hot reload feature, which facilitates rapid UI testing, as well as its extensive package ecosystem support, such as http for API communication, providers for state management, and google\_fonts for aesthetic display customization.

In general, the structure of a mobile application consists of three main layers:

\begin{enumerate}
    \item User Interface Layer – Handles visual display, including text input fields (TextField), send buttons (Send Button), and reply display areas (Chat Bubble).
    \item Logic Layer: Responsible for managing the flow of data communication between users and the API server using the HTTP POST method.
    \item Data Layer: Manages temporary data storage (state management) such as conversation history, response time, and API connection status.
\end{enumerate}

The interaction process in this application takes place through the following flow:

\begin{enumerate}
    \item The user types an Islamic question into the text input field.
    \item The system then sends the question to the server using an HTTP request in JSON format through the API endpoint that was developed in the previous stage.
    \item The API sends the question data to a Reinforcement Learning (Q-Learning)-based chatbot model to be processed and determine the best response.
    \item The answer is returned to the mobile application in JSON format, then displayed on the screen in the form of a chat bubble with a two-way conversation style.
    \item All conversations are temporarily stored in the state memory to maintain the context of the conversation during the session.
\end{enumerate}

The communication architecture between the application and the chatbot model is shown in Figure X, where communication takes place in a client-server format with a request-response format. Flutter acts as a client that sends user input, while the Flask API acts as a server that connects the application to the machine learning model.
To maintain stable communication performance, asynchronous programming methods (async and await functions in Dart) are used to ensure that the UI remains responsive even though the processing on the server takes relatively longer.

In terms of design, the application interface was created using a minimalist Islamic design approach, using a soft color palette such as olive green and ivory white to represent religious nuances and tranquility. Visual elements such as crescent moon icons and simple calligraphy were also added as the branding identity of the \textit{Sharia} chatbot.

\subsection{Evaluation}

The evaluation stage served as a comprehensive process aimed at assessing the overall performance and practicality of the developed chatbot system. This stage was divided into two major perspectives: technical evaluation and user-centered evaluation. Each perspective contributed unique insights that together created a holistic understanding of how effectively the system could function as a digital medium for \textit{Sharia}-based consultation.

From the technical perspective, a series of structured tests were conducted to examine key performance indicators, including response accuracy, system processing speed, and the chatbot’s capability to interpret the context of user inquiries. Response accuracy was measured by analyzing how well the chatbot's answers aligned with the intended meaning of the questions, particularly those that involved paraphrasing or contextual variation. Meanwhile, system speed was evaluated by recording the duration required for the model to receive a question, process it through the embedding and \textit{Q-Learning} mechanisms, and return a response. Consistent response times indicated that the system was efficient and technically stable. Context understanding tests further assessed whether the chatbot could maintain relevance even when users employed diverse sentence structures or indirectly formulated questions.

In addition to technical testing, the user perspective played a crucial role in evaluating the system’s practical usability. This aspect was examined through User Acceptance Testing (UAT), which involved distributing questionnaires to individuals who interacted directly with the chatbot. The questionnaires gathered feedback on several dimensions, including user satisfaction, perceived ease of use, clarity of responses, and the level of trust users felt toward the system's answers. Respondents were also encouraged to provide qualitative input, such as suggestions, critiques, or observations during their interaction with the chatbot. These insights offered valuable information about how real users perceived the chatbot’s utility, reliability, and comfort level in engaging with a digital consultation system for Islamic topics.

The combination of technical performance metrics and user acceptance feedback formed a strong foundation for determining the system’s effectiveness as a digital \textit{Sharia} consultation platform. High technical accuracy alone would not be meaningful without user trust and usability, while positive user experiences must also be supported by stable system performance. Through this dual-layered evaluation, the study was able to identify both the strengths of the system and the areas requiring improvement, ultimately providing a balanced and realistic view of the chatbot’s readiness for broader real-world application.
    
\subsection{Development}
The final stage of the CRISP-DM methodology is deployment, which is the implementation of the system into the operational environment. At this stage, the chatbot that has gone through the testing process is integrated into a digital platform, such as a website or instant messaging application (e.g., Telegram or WhatsApp Bot API), so that it can be widely accessed by users. In addition, a monitoring process is carried out to monitor system performance and regularly update the dataset so that the chatbot remains relevant and able to respond well to new questions. This stage marks the transition from a prototype system to a functional application that is ready for use by the public.

As a further development, this research also designs an \textit{Application Programming Interface} (API) that functions as a bridge between the chatbot model and the \textit{frontend} application used by the users. The API development aims to provide two-way communication services between users and the chatbot system in real-time, so that each question submitted can be immediately processed and answered by the model quickly and efficiently.

The workflow of the API communication can be seen in the following Figure:
\begin{figure}[htb!]
    \centering
    \includegraphics[width=1\linewidth, height=6.5cm]{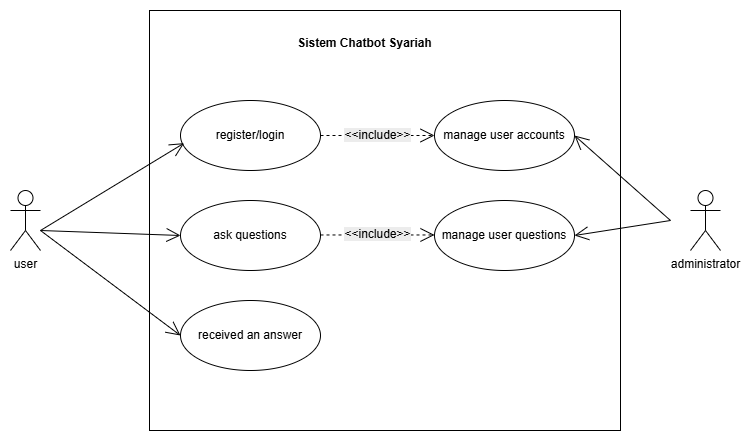}
    \caption{API Communication Flow}
    \label{fig:metod}
\end{figure}
Next, the development of a mobile application prototype based on Flutter was carried out as a user interface (\textit{user interface}) to facilitate the interaction process with the chatbot in a real environment. This application allows users to type questions in text form through the available text field. The question or prompt is then sent to the chatbot API using the HTTP POST method. The API then forwards the data to the chatbot model for processing, and the resulting answer is sent back to the application in the form of a text response displayed on the user’s screen.

Flutter was chosen as the development framework because it has cross-platform capabilities, high development time efficiency, and flexibility in communication with external APIs. Thus, the combination between the API and the Flutter-based mobile application ensures that the \textit{Sharia} chatbot can be directly tested by users through a simple, interactive, and easy-to-understand interface.

\section{Result and Discussion} \label{sec:result}

\subsection{Result}

The results of this study illustrate that the integration of the \textit{Q-Learning} reinforcement learning algorithm with the \textit{Sentence-Transformers} semantic embedding model offers a promising and practical approach for developing a functional \textit{Sharia}-based chatbot system. Throughout the training phase, the model gradually built a comprehensive \textit{Q-Table} that encoded the relationships between question embeddings as system states and the possible answer selections as system actions. As iterative updates were carried out, the \textit{Q-values} steadily converged toward stable optima, signaling that the learning agent had reached a point of equilibrium. At this stage, the model consistently selected responses that exhibited the highest semantic alignment with the given input. This convergence not only reflects the internalization of semantic patterns within the Islamic dataset but also demonstrates the model’s ability to construct a structured decision-making strategy rooted in reinforcement-based optimization.

As the system transitioned to the inference phase, its functional abilities became more evident. In this stage, incoming user queries were encoded into high-dimensional semantic vectors through the Sentence-Transformers model, enabling the chatbot to interpret meaning rather than rely solely on lexical similarity. The encoded vectors were then compared against the dataset embeddings using cosine similarity, a method well-suited for capturing nuanced variations in meaning. This process allowed the system to identify which stored question best matched the intent of the user’s input. What strengthened the decision-making process further was the integration of the learned \textit{Q-values}, which acted as an additional reference for ranking potential responses. Rather than depending exclusively on similarity scores, the system was able to merge semantic proximity with reinforcement-guided preferences, resulting in a more refined and reliable answer selection mechanism.

The deployment process formed another vital stage in demonstrating the practical viability of the proposed system. An API was implemented as the backbone for communication between the trained \textit{Q-Learning} model and the user-facing application. This API handled the full pipeline of receiving user queries, processing them through the embedding and reinforcement logic, and returning system responses in real time. To evaluate this pipeline under conditions closer to real-world usage, a prototype mobile application was developed using Flutter. The application employed a straightforward and intuitive user interface, enabling users to type questions, transmit them seamlessly to the backend model, and instantly view the chatbot’s responses. The mobile prototype not only served as a proof of concept but also as a controlled testing platform where interaction flow, latency, usability, and semantic accuracy could be evaluated comprehensively. Figure 4.1 provides an illustration of this interface, showcasing how the input field, output area, and overall layout were intentionally designed to support efficient functional testing.

The model’s performance during functional testing reaffirmed the effectiveness of the proposed approach. When presented with a wide variety of questions relevant to the Islamic domain, the chatbot demonstrated strong semantic understanding, achieving an overall accuracy of 87\%. This performance level indicates that the system possesses the ability to generalize meaningfully, even when the phrasing or structure of user queries differs from the examples included in the training dataset. These results highlight the strength of combining semantic embeddings with reinforcement learning patterns, particularly in closed-domain environments where precision and contextual reliability are expected. The efficient API handling and the lightweight Flutter application further ensured that responses were delivered quickly, making the system suitable for real-world religious consultation settings.

Despite these promising outcomes, several limitations remain. One key constraint is the static nature of the \textit{Q-Table}; once the model is deployed, it does not continue to learn or update its internal knowledge. This means that newly emerging expressions, terms, or contextual issues in Islamic discourse cannot be accommodated unless the system undergoes retraining. Moreover, the quality of the chatbot's responses is highly dependent on the depth, accuracy, and representativeness of the dataset used. In cases involving complex theological inquiries or multi-layered contextual interpretations, the system may still provide answers that are semantically correct but not entirely aligned with deeper jurisprudential reasoning. These observations suggest that while semantic similarity and reinforcement rules offer a solid foundation, more advanced layers such as context modeling, rule-based refinement, or hybrid reasoning may be needed to handle sophisticated Islamic queries.

In conclusion, the findings of this research demonstrate that integrating \textit{Sentence-Transformers} with \textit{Q-Learning} can produce a functional, semantically-aware chatbot capable of supporting Islamic question-answering tasks. The successful deployment through an API and a mobile prototype further showcases the feasibility of applying this method within digital da’wah and educational environments. With continued improvement particularly in dataset expansion, adaptive learning capabilities, and enhanced contextual interpretation the system holds significant potential to evolve into a more intelligent, adaptive, and reliable Islamic consultation platform that can meet the growing needs of modern users.

\begin{figure}[htb!]
    \centering
    \includegraphics[width=0.5\linewidth]{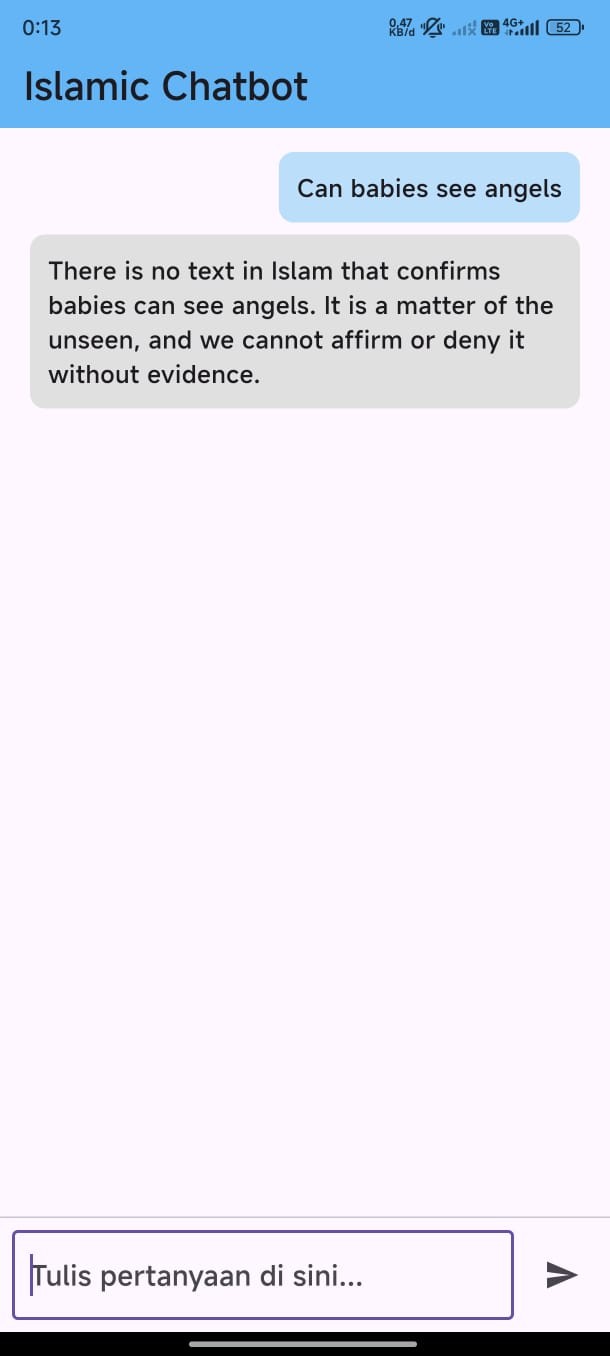}
    \caption{Mobile Application Prompt}
    \label{fig:metod}
\end{figure}

However, this system still has several limitations. Since the model does not dynamically update the Q value after the initial training process is complete, the system does not have the ability to continue learning from new interactions directly. In addition, the quality of the answers is highly dependent on the diversity, depth, and representativeness of the content in the dataset used. If the dataset lacks sufficient coverage of certain Islamic topics, contains incomplete explanations, or exhibits limited variations in question phrasing, the system may generate answers that appear semantically aligned but are not fully contextually accurate. In certain cases, when the question is very specific or multidimensional, the system tends to select an answer that is close in meaning but not necessarily the most appropriate or jurisprudentially precise.

From a methodological perspective, the tabular nature of the Q-Learning algorithm also presents inherent constraints. Q-Learning is not optimized for environments with large or continuous state spaces, making it less scalable when dealing with high-dimensional embeddings generated by Sentence-Transformers. Because the model depends on a static Q-Table, it cannot adaptively update action preferences or improve its decision-making patterns without full retraining. This limitation suggests the potential benefit of more advanced reinforcement learning techniques, such as Deep Q-Networks (DQN), which could handle continuous semantic representations more effectively while enabling ongoing learning.

Furthermore, real-time performance is influenced by computational constraints associated with the embedding process. The generation of sentence embeddings remains relatively intensive, especially when processed on lower-end devices. As a result, system latency is still influenced by server performance, network conditions, and backend loading time. The chatbot relies heavily on API communication between the mobile application and backend server, meaning that disruption in connectivity or insufficient server resources can degrade the user experience. This dependency poses challenges for deployment in environments with limited infrastructure or inconsistent network availability.

Overall, the results of this study show that the combination of Sentence-Transformers and Q-Learning can be used effectively to build complex, semantically-aware chatbots. With improvements to dataset quality, expansion of knowledge sources, incorporation of more advanced reinforcement learning architectures, and the potential addition of continuous learning mechanisms, this system can be further developed to meet broader user needs while ensuring higher accuracy, adaptability, and long-term relevance in Islamic digital consultation contexts.

\subsection{Functional Testing}

Functional testing was conducted as a crucial stage to evaluate the chatbot’s ability to understand, interpret, and produce semantically appropriate responses to new questions that were not included in the training dataset. The primary focus of this evaluation was not merely to determine whether the chatbot could provide an answer, but to assess whether the answer was contextually coherent, relevant, and aligned with validated Islamic knowledge. This testing phase served as a reflection of the model’s generalization capability when faced with variations in linguistic style, question structure, and user expression commonly encountered in real conversational settings.

A total of 100 testing scenarios were designed to represent diverse contexts within the Islamic domain, including worship, jurisprudence, social ethics, \textit{muamalah}, and contemporary issues frequently discussed by users. Each scenario consisted of questions formulated with different phrasing, covering paraphrased queries, implicitly stated questions, and sentences that varied in structure but conveyed similar semantic intent. This approach ensured that the evaluation measured the model’s ability to grasp meaning rather than simply match keywords.

The testing results demonstrated a commendable performance. Out of 100 scenarios, 87 responses were categorized as relevant, 9 as fairly relevant, and only 4 as not relevant. Overall, the system achieved a semantic accuracy rate of 87\%, indicating that the chatbot consistently captured the essence of user questions and delivered responses that remained aligned with the intended context, despite significant variation in sentence construction. In other words, the chatbot exhibited strong generalization capabilities when interacting with different linguistic expressions.

Interestingly, although the model was trained within a closed-domain scope, it was still able to interpret paraphrased and contextually implied questions with notable accuracy. This capability was not accidental; rather, it stemmed from the integration of the Q-Learning algorithm with the Sentence-Transformers model. Q-Learning contributed to optimizing response selection through interactive learning patterns, while Sentence-Transformers provided deep semantic understanding, enabling the system to evaluate meaning beyond surface-level textual similarity.

Overall, the results of the functional testing indicate that the developed chatbot is not only capable of retrieving answers from its predefined knowledge base but also functions effectively as a reliable digital consultation tool. With stable accuracy and strong contextual comprehension, the system demonstrates significant potential for delivering accurate, informed, and trustworthy Islamic information within the boundaries of verified religious knowledge.

\section{Conclusion} \label{sec:conclusion}
This study successfully implemented a \textit{Sharia}-based chatbot system utilizing a combination of the Q-Learning reinforcement learning method and the Sentence-Transformers semantic embedding model. The integration of both approaches enabled the system to understand user questions contextually and generate semantically relevant responses through a structured process of similarity matching. By representing questions in vector space and selecting answers based on the highest semantic similarity score, the chatbot demonstrated the capability to operate effectively within a closed-domain Islamic knowledge environment. The functional testing phase showed that the system achieved a semantic accuracy rate of 87\%, indicating strong generalization performance and reliability in responding to various Islamic queries.

The development of a mobile application prototype using Flutter further proved the practical applicability of the system. Through seamless API communication with a Flask backend, users were able to interact with the chatbot in real time, reinforcing its potential as a digital consultation tool. This implementation highlights how AI can be leveraged to support digital da’wa initiatives, expand access to Islamic knowledge, and enhance religious literacy across different segments of society. The prototype demonstrates that the system can be integrated into broader digital platforms such as educational applications, Islamic information portals, and institutional religious services.

Despite these achievements, several limitations remain. The current system does not support continuous learning after deployment, causing the chatbot to rely solely on its initial dataset without the ability to dynamically adapt to new expressions or emerging issues in contemporary Islamic discourse. In addition, the accuracy of the system is still largely dependent on the completeness, diversity, and representativeness of the dataset. Queries that fall outside the distribution of the original dataset may lead to suboptimal or semantically distant responses. Furthermore, the model has not yet incorporated mechanisms for context retention across multiple-turn conversations, which is essential for handling complex consultations involving follow-up questions.

\begin{figure}[ht]
    \centering
    \includegraphics[width=0.5\linewidth]{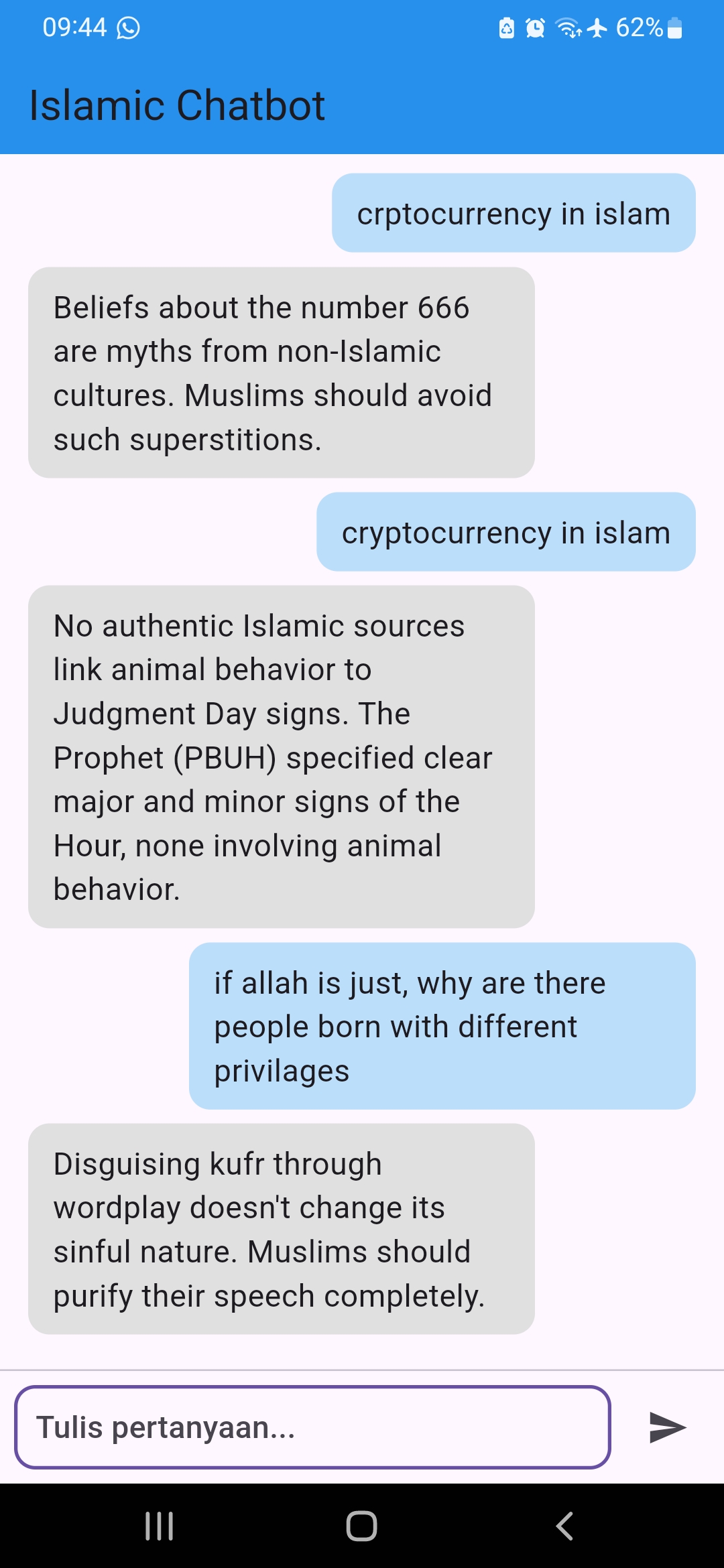}
    \caption{Mobile Application Prompt Result}
    \label{fig:metod}
\end{figure}

To address these limitations, future research directions may include expanding the dataset with more comprehensive and multilingual Islamic sources, integrating Reinforcement Learning from Human Feedback (RLHF) to refine answer selection, incorporating semantic context tracking for multi-round conversations, and enhancing user interaction through speech-based input and output. Additional improvements such as user profiling, adaptive learning, and integration with cloud-based knowledge repositories could also elevate the system’s intelligence and practical utility.

In conclusion, the \textit{Sharia} chatbot developed in this research is not merely a technological prototype but a meaningful step toward the modernization of Islamic knowledge dissemination. It bridges traditional scholarly resources with contemporary digital needs, providing a structured, ethical, and accessible platform for religious consultation. This research lays an important foundation for future innovations in Islamic AI services and contributes to the broader movement of Islamic Digital Transformation. With continued development and responsible adoption, AI-driven consultation systems like this can play a valuable role in empowering communities, supporting religious education, and ensuring that authentic Islamic knowledge remains accessible in an increasingly digital world.

\section*{Acknowledgment}
The author's wishes to acknowledge the Informatics Department \textit{UIN Sunan Gunung Djati Bandung}, which partially supports this research work.

\bibliographystyle{./IEEEtran}
\bibliography{./IEEEabrv,./IEEEkelompok1}

@article{Sihotang2020,
  author = {Sihotang, M. T. and Jaya, I. and Hizriadi, A. and Hardi, S. M.},
  title = {Answering Islamic Questions with a Chatbot using Fuzzy String-Matching Algorithm},
  journal = {Journal of Physics: Conference Series},
  volume = {1566},
  number = {1},
  pages = {012002},
  year = {2020},
  doi = {10.1088/1742-6596/1566/1/012002}
}

@article{Othman2022,
  author = {Othman, A. E. M. and Ewees, A. A. and Alabsi, A. A. and Abdo, M. A.},
  title = {Automated Islamic Jurisprudential Legal Opinions Generation Using Artificial Intelligence},
  journal = {Pertanika Journal of Science \& Technology},
  volume = {30},
  number = {2},
  pages = {689--704},
  year = {2022}
}

@article{Suwarman2025,
  author = {Suwarman, Y. and Syafruddin, A. and Suwena, K. and Sutiah},
  title = {Islamic QA with Chatbot System Using Convolutional Neural Network},
  journal = {International Journal of Advanced Computer Science and Applications},
  volume = {16},
  number = {4},
  year = {2025}
}

@article{Nurhapiza2024,
  author = {Nurhapiza, N. and Harahap, N. S. and Fikry, M. and Affandes, M.},
  title = {Penerapan Chatbot pada Aplikasi Web Tanya Jawab Tentang Fiqih Jual Beli Islam Menggunakan LangChain},
  journal = {Journal of Computer System and Informatics (JoSYC)},
  volume = {5},
  number = {3},
  pages = {548--557},
  year = {2024}
}

@article{Helviansyah2025,
  author = {Helviansyah, T. and Harahap, N. S. and Irsyad, M. and Negara, B. S.},
  title = {SISTEM TANYA JAWAB BERBASIS CHATBOT WEBSITE MENGGUNAKAN GEMINI AI PADA DATA FIQIH KONTEMPORER},
  journal = {Journal of Information System Management (JOISM)},
  volume = {5},
  number = {2},
  year = {2025}
}

@article{Rahayu2024,
  author = {Rahayu, S. and Harahap, N. S. and Agustian, S. and Pizaini, P.},
  title = {Penerapan Teknologi LangChain pada Question Answering System Fikih Empat Madzhab},
  journal = {MALCOM: Indonesian Journal of Machine Learning and Computer Science},
  volume = {4},
  number = {3},
  pages = {974--983},
  year = {2024}
}

@article{Permadi2024,
  author = {Permadi, A. B. and Harahap, N. S. and Handayani, L. and Yusra},
  title = {IMPLEMENTASI QUESTION ANSWERING SYSTEM TAFSIR AL-AZHAR MENGGUNAKAN LANGCHAIN DAN LARGE LANGUAGE MODEL BERBASIS CHATBOT TELEGRAM},
  journal = {Jurnal Teknoif Teknik Informatika Institut Teknologi Padang},
  volume = {12},
  number = {1},
  pages = {62--69},
  year = {2024}
}

@article{Falah2023,
  author = {Falah, M. B. and Putri, N. E.},
  title = {Artificial Intelligence Berbasis Chatbot: Sarana Baru Panduan Hukum Keluarga Digital},
  journal = {Qisthosia "Jurnal Syariah \& Hukum"},
  volume = {4},
  number = {2},
  pages = {126--138},
  year = {2023}
}

@article{Insana2024,
  author = {Insana, Z. and Satriah, L.},
  title = {Etika dan Tantangan Dakwah di Era Kecerdasan Buatan Studi Kasus Penggunaan Chatbot AI untuk Konsultasi Keagamaan},
  journal = {Jurnal Komunikasi Islam (J-KIs)},
  volume = {5},
  number = {2},
  pages = {259--270},
  year = {2024}
}

@misc{Fathir2024,
  author = {Fathir, M. F. S. A. and Handayani, W. and Mustaqim, M.},
  title = {Analisis Faktor-faktor yang Mempengaruhi Kepuasan Penggunaan Chatbot AISYAH BSI Menggunakan Teori UTAUT 2},
  howpublished = {Skripsi Universitas Islam Indonesia},
  year = {2024}
}

@article{awaliyah2022chatbot,
  author = {Tjut Awaliyah and Aries Maesya and Andjar Saputra},
  title = {Aplikasi Chatbot dan Konsultasi Agama Islam Berbasis Android},
  year = {2022},
  url = {https://eprints.unpak.ac.id/id/eprint/7524}
}

@article{hardiyanti2025haji,
  author = {R. Hardiyanti and others},
  title = {Pengembangan Chatbot Konsultasi Haji dan Umrah Berbasis AI},
  year = {2025},
  url = {https://jurnal.unsur.ac.id/semnastekunsur/article/download/5478/3597}
}

@online{sipaham2025sosialisasi,
  title = {Sosialisasi Chatbot Si PAHAM untuk Perkuat Moderasi Beragama},
  year = {2025},
  url = {https://kemenag.malangkota.go.id/showNews?head=sosialisasi-chatbot-si-paham-untuk-perkuat-moderasi-beragama-dan-deteksi-dini-konflik-sosial}
}

@article{masuzzahra2025hana,
  title={HANA: An AI Chatbot for Islamic Jurisprudence on Menstruation using SBERT and TF-IDF},
  author={Masuzzahra, Tsaura Rafah and Umam, Khothibul and Mustofa, Hery and Handayani, Maya Rini},
  journal={Journal of Applied Informatics and Computing},
  volume={9},
  number={3},
  pages={1013--1024},
  year={2025}
}

@online{saudi2025chatbot,
  title = {Perusahaan AI Saudi Luncurkan Chatbot Berbahasa Arab Selaras Nilai Islam},
  year = {2025},
  url = {https://khazanah.republika.co.id/berita/t1laiq320/perusahaan-ai-saudi-luncurkan-chatbot-berbahasa-arab-yang-selaras-nilai-islam}
}

@article{awaliyah2025syariah,
  title = {Pemanfaatan Kecerdasan Buatan dalam Chatbot Tanya Jawab Fikih Syariah},
  year = {2025},
  url = {https://urj.uin-malang.ac.id/index.php/mij/article/download/17330/4603}
}

@article{helviansyah2025kontemporer,
  title = {Sistem Tanya Jawab Berbasis Chatbot untuk Fiqih Kontemporer},
  year = {2025},
  url = {https://jurnal.amikom.ac.id/index.php/joism/article/view/2082}
}

@article{ethical2024guidelines,
  title = {Ethical Guidelines for AI Chatbots in Islamic Context},
  year = {2024},
  url = {https://unimel.edu.my/journal/index.php/JLG/article/download/1895/1484}
}

@article{anggraini2024islamic,
  author = {Anggraini, Ratih and Tursina, Dara and Sarno, Riyanarto},
  title = {Islamic QA with Chatbot System Using Convolutional Neural Network},
  journal = {Iraqi Journal of Science},
  year = {2024},
  volume = {65},
  number = {4},
  pages = {2232--2241},
  doi = {10.24996/ijs.2024.65.4.38}
}

@article{campbell2020authority,
  author = {Campbell, Heidi A.},
  title = {Religion and the Digital Age: Understanding Religious Authority Online},
  journal = {Social Compass},
  volume = {67},
  number = {2},
  pages = {256--270},
  year = {2020}
}

@article{solomon2024rule,
  title={Rule based chatbot design methods: A review},
  author={Solomon, Elsabeth and Tilahun, Surafel L},
  journal={Journal of Computational Science and Data Analytics},
  volume={1},
  number={01},
  pages={75--84},
  year={2024}
}

@article{okonkwo2021chatbots,
  author = {Okonkwo, Chidera W. and Ade-Ibijola, Abejide},
  title = {Chatbots in Education: A Systematic Review},
  journal = {International Journal of Educational Technology in Higher Education},
  year = {2021},
  doi = {10.1186/s41239-021-00262-4}
}

@inproceedings{zulfa2025peran,
  title={PERAN CHATBOT BERBASIS ARTIFICIAL INTELLIGENCE DALAM MENINGKATKAN KINERJA EKONOMI SYARIAH: ANALISIS DARI PERSPEKTIF MAQASHID SYARIAH},
  author={Zulfa, Siti Launa and others},
  booktitle={Gunung Djati Conference Series},
  volume={56},
  pages={1459--1465},
  year={2025}
}

@article{uriawan2024real,
  title={Real-Time Chatbot: Microservices Implementation in Distributed System Architecture},
  author={Uriawan, Wisnu and Herdiyanto, Reza Fahlevi and Millah, Rd Imam Saepul and Irhamnillah, Sami and Gunawan, Siti Nurhayati},
  year={2024},
  publisher={Preprints}
}

@inproceedings{hardiyanti2025pengembangan,
  title={Pengembangan Chatbot Konsultasi Haji dan Umrah Berbasis RAG dan Prompt Engineering: Studi Kasus Fillah Guide},
  author={Hardiyanti, Rijki and Sany, Diny Syarifah},
  booktitle={PROSIDING SEMINAR NASIONAL TEKNIK UNIVERSITAS SURYAKANCANA},
  volume={2},
  number={1},
  pages={329--337},
  year={2025}
}

@article{sofyan2025implementation,
  title={Implementation of Natural Language Processing (NLP) in Developing a Chatbot Application for Classical Islamic Text Learning at Pesantren El-Huda El-Islamy},
  author={Sofyan, Yan and Arroyan, Alwi Fachri Ibnu},
  journal={Journal TIFDA (Technology Information and Data Analytic)},
  volume={2},
  number={1},
  pages={34--41},
  year={2025}
}

@article{hartawan2024bidirectional,
  title={Bidirectional and Auto-Regressive Transformer (BART) for Indonesian Abstractive Text Summarization},
  author={Hartawan, Gaduh and Maylawati, Dian Sa'adillah and Uriawan, Wisnu},
  journal={Jurnal Informatika Polinema},
  volume={10},
  number={4},
  pages={535--542},
  year={2024}
}

@misc{datasciencePM,
  title={CRISP-DM: Data Mining Methodology},
  author={DataScience-PM},
  year={2020},
  note={Available at: \url{https://www.datascience-pm.com/crisp-dm-2/}}
}

@misc{getsmarterdataMining,
  author = {{getSmarter}},
  title  = {How Artificial Neural Networks Can Be Used for Data Mining},
  year   = {2025},
  note   = {Available at: \url{https://www.getsmarter.com/blog/how-artificial-neural-networks-can-be-used-for-data-mining/} (accessed: 2025-11-19)}
}


\end{document}